\documentclass[runningheads]{llncs}
\usepackage{graphicx}
\usepackage{subcaption}
\usepackage{graphicx}
\usepackage{todonotes}
\usepackage{svg}
\usepackage{amsmath,amssymb}
\DeclareMathOperator{\E}{\mathbb{E}}
\usepackage{mathtools}
\usepackage{bm}
\usepackage{soul}

\begin{document}
\title{Symmetry and Complexity in Object-Centric Deep Active Inference Models}
%

%
\author{Stefano Ferraro \and
 Toon Van de Maele \and
 Tim Verbelen \and 
 Bart Dhoedt}%
\authorrunning{S. Ferraro et al.}
%
\institute{IDLab, Department of Information Technology \\
 Ghent University - imec \\ 
 Ghent, Belgium \\
 \email{stefano.ferraro@ugent.be}
}
\maketitle              
\begin{abstract}

Humans perceive and interact with hundreds of objects every day. In doing so, they need to employ mental models of these objects and often exploit symmetries in the object's shape and appearance in order to learn generalizable and transferable skills. 
Active inference is a first principles approach to understanding and modeling sentient agents. It states that agents entertain a generative model of their environment, and learn and act by minimizing an upper bound on their surprisal, i.e. their Free Energy. The Free Energy decomposes into an accuracy and complexity term, meaning that agents favor the least complex model, that can accurately explain their sensory observations.

In this paper, we investigate how inherent symmetries of particular objects also emerge as symmetries in the latent state space of the generative model learnt under deep active inference. In particular, we focus on object-centric representations, which are trained from pixels to predict novel object views as the agent moves its viewpoint. First, we investigate the relation between model complexity and symmetry exploitation in the state space. Second, we do a principal component analysis to demonstrate how the model encodes the principal axis of symmetry of the object in the latent space. Finally, we also demonstrate how more symmetrical representations can be exploited for better generalization in the context of manipulation.  

\end{abstract}

\section{Introduction}

Humans perceive and interact with hundreds of objects every day. In doing so, they need to develop mental models of these objects, which represent the object shape, color, affordances, etc. From early infancy, toddlers learn this by actively engaging with objects, which makes that they effectively mainly sample the world with close-up observations of objects in center view, often holding and manipulating the object in their hands~\cite{smith2018,slone2019}.

\begin{figure}[t!]
  \centering
  \includegraphics[width=\textwidth/2]{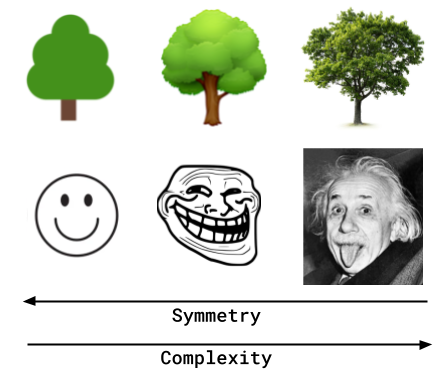}
  \caption{In the brain concept can be represented with different levels of complexity by exploiting different levels of symmetry.}
  \label{fig:intro}
\end{figure}

When representing objects in our mind, we often generate simpler representations than their real-world observations. These representations often exploit the various symmetries that are ubiquitous in real-world objects. For example, if we are asked to quickly sketch a tree or a face we would end up drawing something really similar to the ones presented at the left of Figure~\ref{fig:intro}. Intuitively, this shows how more symmetries arise in the representation as it becomes less complex.

Active inference is a first principles approach to understanding and modeling sentient agents~\cite{friston_history_2012}. It states that agents entertain a generative model of their environment, and learn and act by minimizing an upper bound on their surprisal, i.e. their Free Energy~\cite{friston_active_2016}. Free Energy decomposes into an accuracy and complexity term, meaning that agents favor the least complex model, that can accurately explain their sensory observations.

Recently, deep active inference models have been proposed, which parameterize the generative model using deep neural networks and learn the state space structure using stochastic gradient descent on the variational free energy loss function~\cite{catal_learning_2020,mazzaglia_entropy}. In this paper, we investigate if symmetries also emerge in the latent state space of the generative model learnt under deep active inference and whether this is related to lower complexity in the Free Energy. In particular, we focus on object-centric representations, which are trained from pixels to predict novel object views as the agent moves its viewpoint~\cite{van_de_maele_ccn_2022}.

In summary, we will zoom in on the following aspects:
\begin{enumerate}
    \item We investigate the relation between model complexity and symmetry exploitation in the state space, by quantifying and comparing both on different object-centric models.
    \item We do a principal component analysis to demonstrate how the model encodes the principal axis of symmetry of the object in the latent space.
    \item We demonstrate how more symmetrical representations can be exploited for better generalization of action selection in the context of manipulation.  
\end{enumerate}

In doing so, we will first formally define the concept of symmetries using group theory in Section 2. Next in Section 3, we summarize active inference, variational Free Energy, and how this decomposes into model complexity and accuracy. Section 4 then presents the object-centric deep active inference setup used in this paper, after which we address the aforementioned items in the experiments in Section 5. Finally, we end with a discussion in Section 6, and conclusion in Section 7.

\section{What is symmetry?}
\label{sec:symmetry}

Symmetries are a set of transformations that, when applied to an object, render the object invariant. Mathematically, this can be formalized using group theory~\cite{bronstein2021,higgins_symmety_2022}. A \textit{group} $G$ is defined as a set equipped with a composition operation $G \times G \rightarrow G$, i.e. ``$\cdot$'', that, given two elements $g_1$ and $g_2$ of the set, produces a third element of the set $g_3$:

\begin{flalign}
    &(g_1, g_2)  \rightarrow g_1 \cdot g_2 = g_3, \\ \nonumber
    & g_1, g_2, g_3 \in G
\label{eq:group}
\end{flalign}

given that:
\begin{enumerate}
    \item $g_1 \cdot g_2$ is an associative operation
    \item there exist an identity element $e \in G$ such that $e \cdot g = g$, $\forall g \in G$
    \item for any $g \in G$ there exist a $g^{-1} \in G$ such that $g \cdot g^{-1} = g^{-1} \cdot g = e$
\end{enumerate}

Given a group $G$ and a set of objects $X$, a \textit{group action} of $G$ on $X$ is a mapping $G \times X \rightarrow X$, i.e. an action $g$ of $G$ on $x_1$ of $X$ produces an element $x_2$ of the set $X$:

\begin{flalign}
    &(g, x_1)  \rightarrow g \cdot x_1 = x_2 \\ \nonumber
    &x_1, x_2 \in X 
\label{eq:action}
\end{flalign}

such that:
\begin{enumerate}
    \item $g \cdot x_1$ is an associative operation
    \item there exist an identity element $e \in G$ such that $e \cdot x = x$, $\forall x \in X$
\end{enumerate}

We can now define the concept of an invariant map. Given again a group $G$ that acts on a set $X$, if $F \colon X \rightarrow Y$ is a map between sets $X$ and $Y$, $F$ is invariant if $F(g \cdot x) = F(x), \forall (g, x) \in G \times X$. If $X$ is invariant under such a mapping, the mapping is a \textit{symmetry} of $X$. The group of all transformations under which the object $X$ is invariant is called the \textit{symmetry group} of $X$.

For example, the set of rotations of 0$^{\circ}$, +90$^{\circ}$ and -90$^{\circ}$ forms a group and can be applied to a set of 2D shapes, which is then the group action. Consider $X$ to be the set of all square shapes, then these rotations are symmetries, but not for the set of triangles.

In addition to invariance, we can also define equivariance. Given a group $G$ acting on both spaces $X$ and $Y$ and given a map $H \colon X \rightarrow Y$, $X$ is to be considered equivariant if for any $g \in G$ and any $x \in X$ we have $H(g \cdot x) = g \cdot H(x)$. Simply put it does not matter the order we apply our group transformations. If we consider now an equivariant map $H$ and an invariant map $F$, then we will have $F(H(g \cdot x)) = F((g \cdot H(x))) = F(H(x))$ that is again an invariant mapping. 

\section{Symmetry and Complexity in Active Inference}

Active inference provides a framework to describe the behavior of sentient agents acting in a dynamic environment. This theory postulates that a sentient agent entails a generative model of the environment, and the agent acts in the environment with the imperative of minimizing an upper bound on surprisal, i.e. free energy~\cite{friston_active_2016}.

The generative model comprises the joint probability $P(o_{0:T}, a_{0:T-1}, s_{0:T})$ over sequences of observations $o_{0:T}$, actions $a_{0:T-1}$ and hidden or latent state $s_{0:T}$ over some time horizon $T$. Modeling this as a partially observable Markov decision process (POMDP) as depicted in Figure~\ref{fig:gen_model}.b, this joint probability can be factorized as:

\begin{equation}
    P(o_{0:T}, a_{0:T-1}, s_{0:T}) = P(s_0)P(o_0|s_0) \prod_{t=1}^{T} 
    P(a_{t-1})
    \underbrace{P(s_{t}|s_{t-1}, a_{t-1})}_{\text{Transition model}}\underbrace{P(o_{t}|s_{t})}_{\text{Likelihood model}} \nonumber
\end{equation}

In order to infer beliefs about the hidden state variables, an active agent needs to calculate a posterior belief provided an observation $o_t$, which is in general intractable. To solve this problem the agent resorts to variational inference, which approximates the true posterior $P(s_t|o_t)$ with an approximate posterior distribution $Q(s_t|o_t)$. The objective is then to maximize the evidence lower bound (ELBO)~\cite{kingma_auto-encoding_2014}, or equivalently, to minimize variational Free Energy~\cite{friston_active_2016}:

\begin{equation}
\begin{split}
\mathcal{F}
={}&\sum_t \underbrace{- \log{P(o_{t})}}_{\text{Evidence}} + \underbrace{D_{KL}[Q(s_{t}|o_{t}) || P(s_{t}|o_{t}, s_{t-1}, a_{t-1})]}_{\text{Divergence}}
\\
={}&\sum_t \underbrace{D_{KL}[Q(s_{t}|o_{t}) || P(s_{t}|s_{t-1}, a_{t-1})]}_{\text{Complexity}} - \underbrace{\E_{Q(s_{t}|o_{t})}[\log{P(o_{t}|s_{t})}]}_{\text{Accuracy}}
\end{split}
\label{eq:free_energy}
\end{equation}

Here, the first line shows that $\mathcal{F}$ is indeed an upper bound on the (negative) log evidence, bounded by the KL divergence between the approximate and the true posterior. The second line rewrites the variational Free Energy as a complexity term and an accuracy term. Here, the complexity term is the divergence between the variational density (i.e. posterior belief) and prior beliefs about hidden states. In other words, the agent prefers the least complex model that yields accurate explanations.

In active inference, agents minimize Free Energy either by updating their internal model or by acting on the environment. Actions are selected which are expected to reduce Free Energy in the future. As agents can rely on their generative model to obtain expected observations in the future, they can estimate the expected Free Energy $\mathcal{G}$ for all considered actions~\cite{catal_learning_2020}:

\begin{equation}
\label{eq:expected_free_energy}
\mathcal{G}(a_t) = \underbrace{D_{KL}\big{[}Q(s_{t+1} | a_t) || P(s_{t+1})\big{]}}_{\text{Risk}} + \underbrace{\mathbb{E}_{Q(s_{t+1})}\big{[}H(P(o_{t+1} | s_{t+1}))\big{]}}_{\text{Ambiguity}}
\end{equation}

Here, we used that the agent has prior preferences over future states $P(s_{t+1})$, and hence searches for actions that bring it closer to preferred states while minimizing the expected ambiguity. Actions are then selected by sampling from $P(a_t) = \sigma(-\gamma \mathcal{G}(a_t))$ with $\sigma$ the softmax function and $\gamma$ the temperature.

In deep active inference, the state space structure is not specified upfront but rather learned from data by instantiating the likelihood model, transition model, and approximate posterior by deep neural networks, and performing gradient descent on the variational Free Energy~\cite{catal_learning_2020}. Crucially, this enables the agent to determine the latent state space structure, and potentially exploit symmetries. To this end, the posterior model can be seen as a map from observations to latent state space, and symmetries can arise from transformations in observation space that render the latent state invariant. Moreover, we hypothesize that the complexity term in the variational Free Energy minimization will encourage the model to exploit symmetries in the state space, especially when there are symmetries in the environment that the agent has to model.


\section{Object-centric generative models}

\begin{figure}[t!]
  \centering
  \subfloat[]{\includegraphics[width=\textwidth / 2]{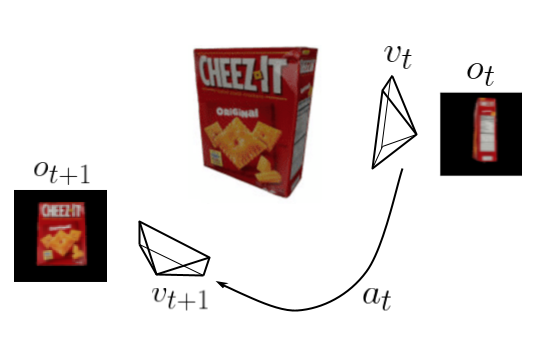}}
  \hspace{2em}
  \subfloat[]{\includegraphics[width=\textwidth / 3]{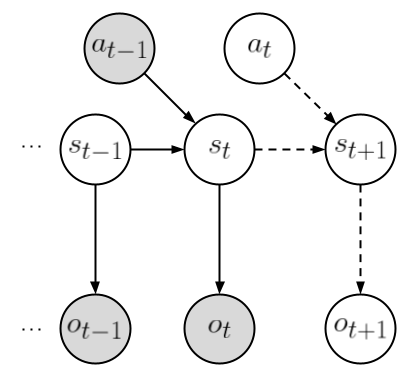}}
  \caption{Visualization of the generative model adopted. (a) the object-centric model adopted, an agent in the form of a pinhole camera is free of moving around the environment while maintaining a focus at the center of mass of the object. Agent move from viewpoint $v_t$ to $v_{t+1}$ through action $a_t$, at each step observation $o$  is rendered. (b) Bayesian Network describing the generative model of the agent. Observation $o_t$ is the output of the generative process steaming from state $s_t$. Latent state $s_t$ is dependent on the previous state $s_{t-1}$ and action $a_{t+1}$. observed variables are presented in grey while unobserved ones are in white.}
  \label{fig:gen_model}
\end{figure}

In order to investigate the level of symmetry that can be exploited through learned generative models, we focus on an agent learning object-centric representations as proposed by Van de Maele et al.~\cite{van_de_maele_ccn_2022}. This model considers observations from different viewpoints of a given object and learns a generative model by minimizing the predicting error for novel views. The agent can move the camera, simulating a robotic manipulator with an in-hand camera, allowing it to sample informative views about the object, and move towards preferred poses, for example where the agent is able to grasp the object. 

This model is effectively implementing a part of a hierarchical generative model of human vision~\cite{parr_generative_2021}, which entails that the brain uses retinal activations to make inferences about the physical scene composition, i.e. which objects with a distinct shape and appearance are present in the environment. Moreover, learning distinct models for separate object categories by looking at objects from a close distance is inspired by the hypothesized role of distinct cortical columns in the brain, which ``vote'' for objects at a certain pose in order to infer a consistent scene representation~\cite{Hawkins2019}. In this work, we particularly focus on these object-centric generative models, as they enable us to investigate the symmetries learnt by the model, and intuitively relate those to the symmetric properties of the physical object at hand that is modeled.

Figure \ref{fig:gen_model} shows the object-centric active inference agent setup, which consists of a pinhole camera that renders views of a particular object. At every timestep $t$, the camera moves to viewpoint $v_t$ and produces an observation $o_t$. Transition to viewpoint $v_{t+1}$ is the result of action $a_t$. This action represents the relative translation and orientation that is applied to the camera pose to acquire a new viewpoint. The action space is thus defined as the collection of transforms that change the camera pose to a different object-centric observation. The resulting observations are images that render the object from various poses, similar to how toddlers focus on particular objects from a close distance in their early infancy~\cite{Smith2011NotYM}. 

This way, learning a latent representation of a particular object is cast as active inference, where the generative model is implemented using deep neural networks. This results in a model consisting of three distinct neural networks: 
\begin{enumerate}
    \item an encoding network $q_{\phi}$ as approximate posterior $Q(s_t | o_t)$ ;
    \item a transition network $p_{\chi}$ as transition model $P(s_{t+1} | s_t, a_t)$;
    \item a decoding network $p_{\psi}$ as likelihood model $P(o_t | s_t)$.
\end{enumerate}

\begin{figure}[t!]
  \centering
  \includegraphics[width=\textwidth]{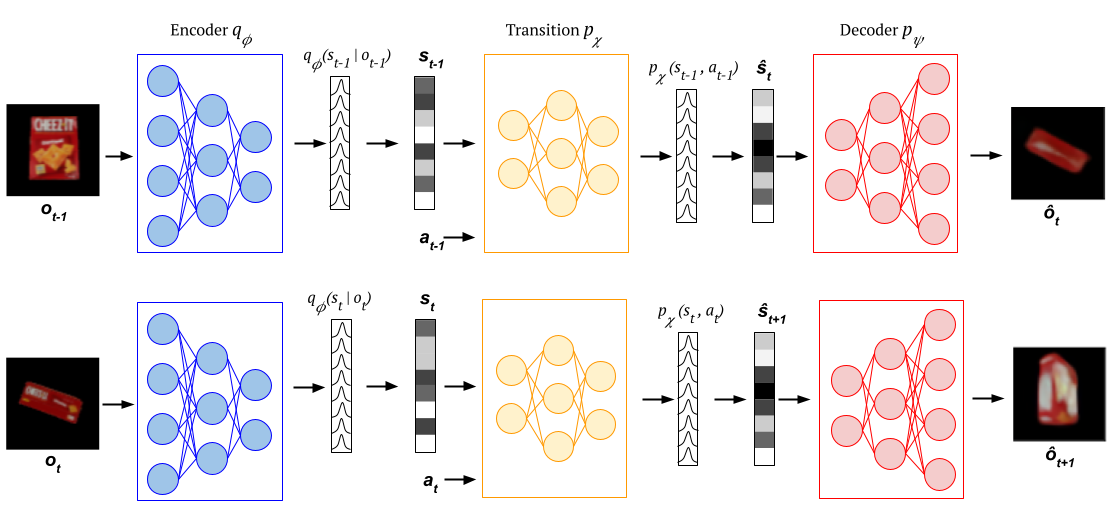}
  \caption{Neural network architecture. From the top left: observation $o_{t-1}$ is fed through the encoding unit $q_{\phi}$, and the output is a belief over the state of the object. Sampling is performed on the latter distribution to obtain a state $s_{t-1}$, which is paired with action $a_{t-1}$ and fed to the transition unit $p_{\chi}$. action $a_{t-1}$ is a relative pose transformation between viewpoint $v_{t-1}$ and $v_t$. The produced belief over the transitioned state undergoes sampling to obtain prediction $\hat{s_t}$.}    
  \label{fig:model_arch}
\end{figure}

A graphical representation is presented in Figure \ref{fig:model_arch}. At timestep $t-1$, observation $o_{t-1}$ is encoded. The result of the encoding process is a belief distribution $q_{\phi}(s_{t-1} | o_{t-1})$, of which a latent state $s_{t-1}$ is sampled using the reparameterization trick. The sampled latent, paired with action vector $a_{t-1}$ is the input of the transition network $p_{\chi}$. The action vector consists of a relative translation and rotation as a 6 DOF continuous rotation representation \cite{Zhou_2019_CVPR}. A belief over the transitioned state is encoded in $p_{\chi}(s_{t-1}, a_{t-1})$. Again sampling will provide the expect transitioned state $\hat{s}_t$. The latter is then decoded by network $p_\psi$ to result in the sensory prediction $\hat{o}_t$. All the networks are jointly trained by minimizing the variational Free Energy:
\begin{equation}
    \label{eq:free_energy_beta}
    \mathcal{F}_{t} = ||o_t - \hat{o}_t||^2 + \beta D_{KL}[Q(s_{t}|o_{t}) || P(s_{t}|s_{t-1}, a_{t-1})]  
\end{equation}

The first term is the accuracy which is computed as the mean squared error between the prediction $\hat{o}_t$ and the ground truth $o_t$. The second term is the complexity term which we scale with a $\beta$ coefficient to adapt the weight of the model complexity. The resulting loss formulation is hence similar as proposed by the $\beta$-VAE \cite{higgins2017betavae}. By varying $\beta$ we will assess to what extent model complexity relates to emerging symmetries in the latent state space. Given the formulation similarities with the $\beta$-VAE architecture, we expect the information bottleneck imposed by the variation of $\beta$ will result in a compression of the latent space and further exploitation of symmetric features.

\section{Experiments}

\begin{figure}[t!]
  \centering
  \includegraphics[width=\textwidth]{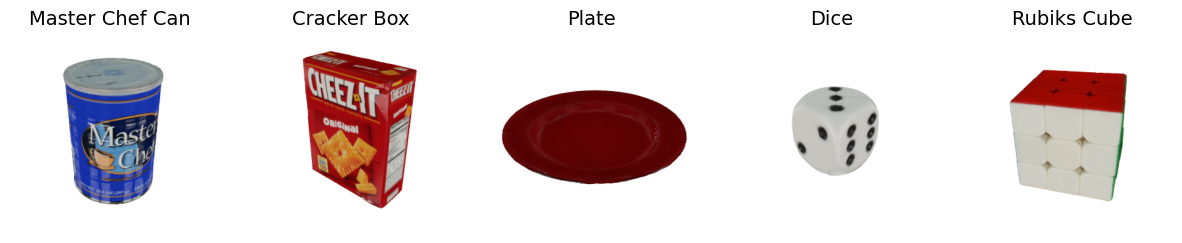}
  \caption{Subset of YCB dataset considered for the experimentations.}
  \label{fig:objects}
\end{figure}

We trained the presented model on a subset of the YCB dataset~\cite{xiang_posecnn_2018}.
Objects are shown in Figure \ref{fig:objects}. They have been selected based on the presence of axes of symmetry both for the texture and shape of the object (i.e. Master Chef can and plate ) and the presence of only shape symmetry (i.e. cracker box, dice, and Rubik's cube). 

For each object, a set of 10000 randomly sampled observations and their corresponding viewpoints have been recorded. These observations are produced in an object-centric fashion, by sampling the azimuth and elevation from a spherical coordinate system, while the radius is kept at a fixed value.

We use the model architecture illustrated in Figure \ref{fig:model_arch}. The training procedure is similar to the one adopted by Van de Maele et al.~\cite{van_de_maele_ccn_2022} and consists of processing pairs of observation simultaneously.

For each object, we trained different models with $\beta$ varying in a range between 0.25 to 100. In the following subsections, we will analyze the resulting latent space structure. First, we evaluate how the resulting model complexity changes with varying $\beta$, and how this results in invariants in the state space which reflects the symmetry. Next, we do a principal component analysis and check whether the amount of principal components relate to the symmetry axes of the modeled object. Finally, we show how exploiting symmetry can help to generalize action selection, i.e. to infer successful grasp poses from a demonstration.

\subsection{Complexity and Symmetry exploitation}

In order to evaluate model complexity, we consider a set of 900 evaluation pairs consisting of two randomly sampled observations and the action moving the camera viewpoint from one to the other. For each pair, we calculate the complexity as the KL divergence between the posterior and the prior. The prior is obtained by encoding the first observation with the posterior model, and then predicting the state distribution using the transition model given the action. The posterior distribution is calculated by encoding the second observation with the posterior model. We evaluate the complexity term for the whole evaluation dataset and report the median as an overall measure of complexity. 

\begin{figure}[t!]
  \centering
  \includegraphics[width=\textwidth]{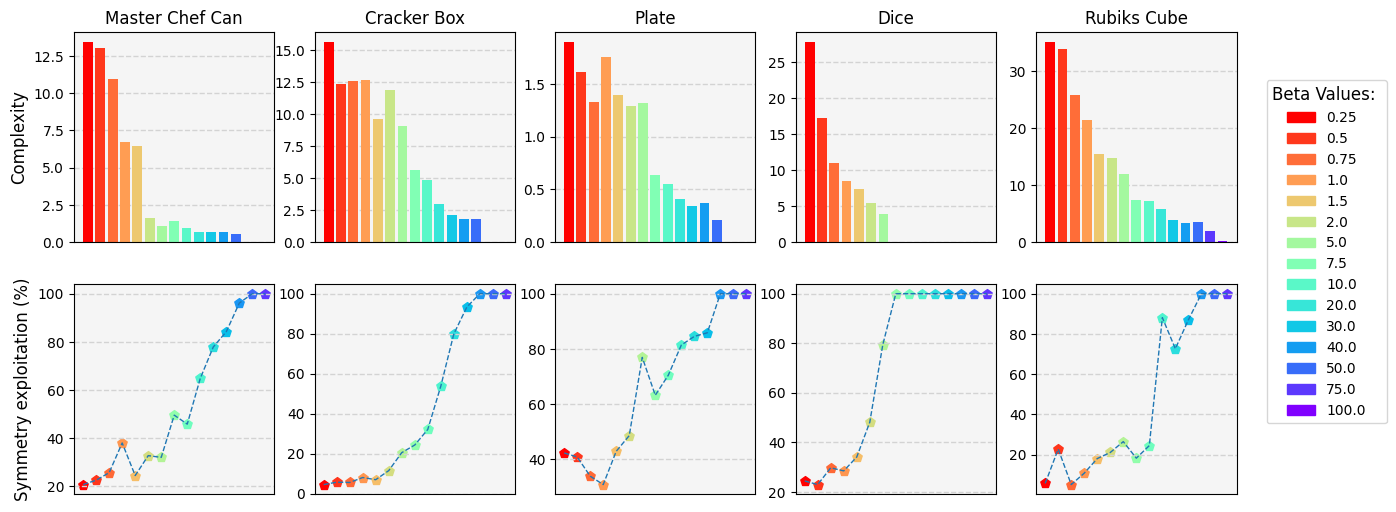}
  \caption{Complexity terms and symmetry exploitation. Evaluated on a subset of the YCB dataset. For each object, multiple models have been trained adjusting the beta factor. The symmetry exploitation term is expressed in percentage terms.}
  \label{fig:complexity}
\end{figure}

To evaluate the symmetry exploitation from the model, we analyze to what extent the latent state is invariant for different object viewpoints. To this end, we use the same evaluation dataset, but now encode both observations with the posterior model, and measure the KL divergence between those two. For a single pair of observations $a$ and $b$ the symmetry term is:
\begin{equation}
    \label{eq:symmetry_exploitation}
    \mathcal \mathcal{S}_{a-b} = D_{KL}[Q(s_{a}|o_{a}) || Q(s_{b}|o_{b}) ]  
\end{equation}

We consider the state mapping invariant when $\mathcal{S}$ is below a certain threshold, which we empirically set to 300. The symmetry exploitation term expresses the percentage of objects' observations that are considered symmetric. 

Our results are shown on Figure \ref{fig:complexity}. The top row shows the model complexity for various objects and varying $\beta$ factors. As expected, model complexity decreases as $\beta$ increases. Conversely, symmetry exploitation increases with increasing $\beta$, until the model collapses for very high $\beta$. In this case, every observation is mapped to a single latent, and we get posterior collapse.

\begin{figure}[t!]
  \centering
  \includegraphics[width=\textwidth]{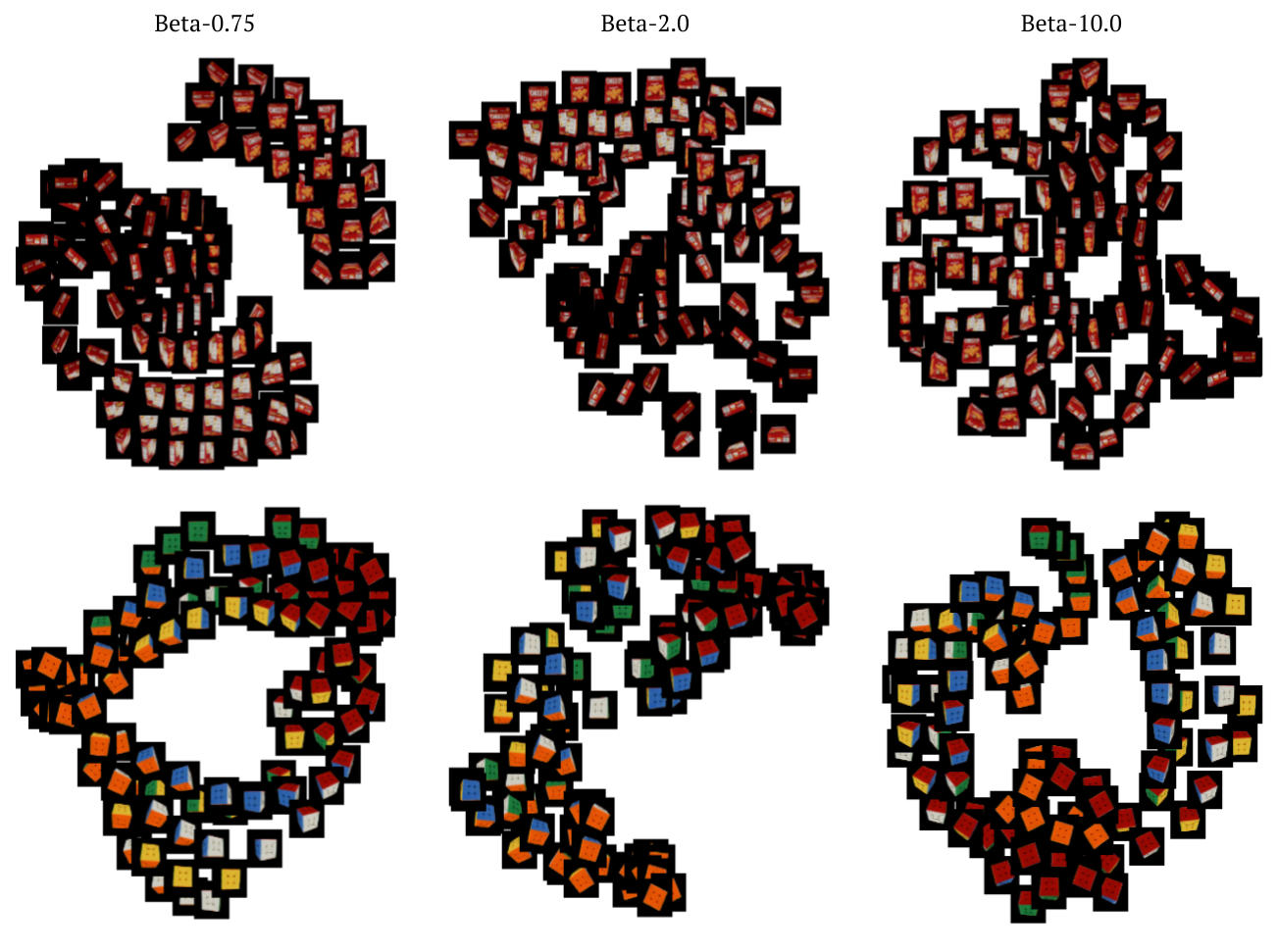}
  \caption{Visualization of the t-SNE mapping applied on the latent space for 100 uniformly distributed observations of cracker box and Rubik's cube. Increasing beta values are presented.}
  \label{fig:t-sne}
\end{figure}

In Figure \ref{fig:t-sne} we provide a visualization of the state space distribution obtained by applying t-SNE \cite{vanDerMaaten2008}. Notice how for a low beta value of 0.75 similar observations are encoded to unique regions of the embedding, with evident separation from dissimilar observations. By increasing beta, we start to notice some degradation in the quality of the partitioning. The shape properties of the objects are still encoded correctly, unlikely the texture ones. For example, for the cracker box, the front and the back of the box are mixed together, instead for the Rubik's cube orange faces are mixed with the red ones. 
Overall, we find that minimizing model complexity results in invariants in the latent space, effectively mapping different observations to the same latent representation. The observations that get mapped to the same latent code also reflect the symmetries of the physical object.

\subsection{Principal Axes of Symmetry}

In this section, we further investigate how the true axes of symmetry of the objects are represented in the latent space. To perform such analysis we opted for Principal Components Analysis (PCA).
PCA is a technique often used for dimensionality reduction \cite{wauthier2020,KEERTHIVASAN2016510}, the output of such analysis is the eigenvectors and eigenvalues of the input dataset. We composed a dataset with 900 rendered observations, obtained by sampling viewpoints with varying elevation and azimuth coordinates, but at a fixed range. 
All observations are encoded into latent states using the posterior model, which we stack into a matrix $M_l$.

\begin{figure}[t!]
  \centering
  \includegraphics[width=\textwidth]{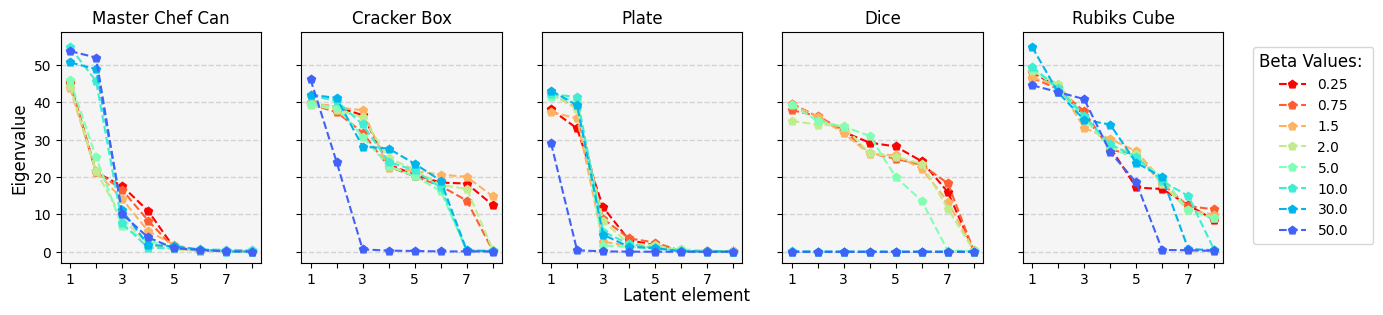}
  \caption{Eigenvalues plots resulting from Principal Component Analysis (PCA). Provided a dataset of 900 instances obtained from the variation along azimuth and elevation with respect to spherical coordinates. PCA is applied to the latent vectors encoded from each observation.}
  \label{fig:pca}
\end{figure}

We hypothesize that for objects that have physical axes of symmetry along the elevation and azimuth dimensions, only two principal components would result from the analysis of the components on $M_l$. Figure \ref{fig:pca} shows the eigenvalues for various objects and varying $\beta$. Indeed, for objects like the Master Chef can and the plate, which have strong symmetry, there are only two components that have large eigenvalues. For objects where the symmetrical properties are not exploited during the data collection, the trend is reflected by a more uniform distribution of the information along all the latent elements. 

Note that once the model collapses for high $\beta$ values, most of the eigenvalues become zero, since no more elements are any more relevant for the collapsed representation. Also, for Master Chef can and the plate it is interesting to point out how for increasing values of $\beta$ the principal axes become more apparent.

\subsection{Generalizing graspable poses through symmetry exploitation}

In this final experiment, we look at how the exploitation of symmetry in the agent's generative model can influence a practical task. To this end, we provide the agent with the task to grasp an object, given an example observation. This preferred observed view can be interpreted as a proxy for the target grasping pose, i.e. as if the agent is a robotic manipulator with an in-hand camera. 

To solve this task, the agent has to plan its action that will realize the preferred observation. The active inference agent does this through the minimization of expected free energy as described in Section~\ref{sec:symmetry}. The considered preference is thus provided as an RGB image $o_{t+1}$ from a randomly sampled viewpoint $v_{t+1}$. The agent's task is then to infer the correct action to move from initial observation $o_t$ to the preference. For this experiment, again a dataset is created with two observations, and the goal is to find an action $a_t$, that brings the agent from the first observation to an observation similar to the second. 

For finding the action with the lowest expected free energy, a Monte Carlo procedure is used. First, the approximate posterior over state $Q(s_t|o_t)$ is computed by encoding observation $O_t$ using $q_\phi$. Then 900 actions are randomly sampled to yield a uniform distribution of target viewpoints around the object. For each of these actions, the transitioned belief is acquired through $p_\chi$. The expected free energy G is then evaluated by computing the negative log probability of the mean of the transitioned state belief with respect to the approximate posterior over the state, given our preferred view $Q(s_{t+1}|o_{t+1})$. We repeat this process for all elements in the dataset.

\begin{figure}[t!]
  \centering
  \includegraphics[width=\textwidth]{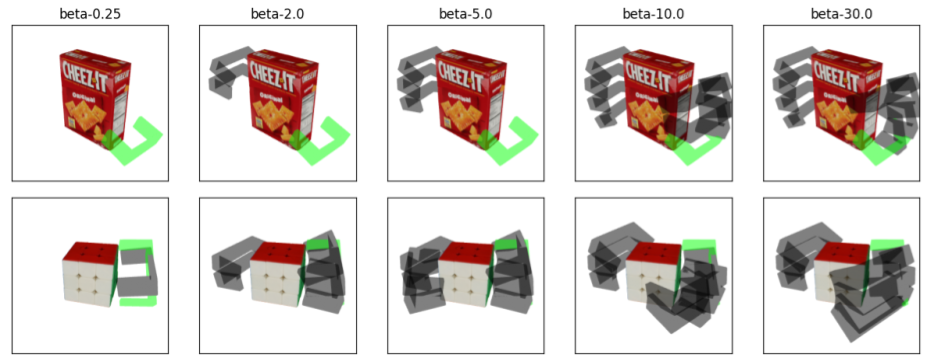}
  \caption{Manipulative actions driven by expected free energy action selection. Provided a target pose, displayed in green, possible selected manipulation poses are shown in black.}
  \label{fig:action}
\end{figure}

We extract the ten actions with the lowest expected free energy and visualize them in Fig~\ref{fig:action} as the corresponding grasping poses. The target pose, or preference, is shown in green while the selected poses are presented in dark gray. From this visualization, it can be observed that by increasing the $\beta$ parameter, the model starts exploiting more symmetries in the system. The relaxation of $\beta$ during the optimization of the model yields behavior where the agent is more permissive towards different -but similar- poses. While this also causes a detrimental effect on reconstruction accuracy, one can see how finding more equally graspable poses for an object is advantageous. As a result, enforcing this exploitation of symmetry aids in the generalization towards other graspable poses. If the manipulation task requires high precision, we can adopt a model with a low $\beta$ and thus low variance over sampled poses, while in the opposite case where we don't care about the exact grasping location, a model with high $\beta$ could be beneficial. 

\section{Discussion}

\subsection{Complexity-accuracy trade-off in view of preferences}

From our experiment results, we conclude that penalizing the complexity of the model through the weighing of the complexity term in the free energy formulation yields an increased amount of symmetry representations. The model has learned different observations mapping to the same belief over the state. In our case, we adapted the model complexity by varying weight $\beta$ on the complexity term in the loss function. In reality, however, the accuracy-complexity trade-off should be governed by preferences. High precision preferences on particular outcomes might require a more complex model (i.e. I want to grasp the object from one specific viewpoint), whereas lower precision preferences (i.e. I just want to grasp the object) can be realized with a less complex model, i.e. one that exploits symmetries.

In an analogous way, Battaglia et al. \cite{battaglia2013} studied the ``intuitive physics'' we apply to the objects we observe or interact with. Battaglia et el. conclude that humans perform this type of inference about object dynamics in an approximate manner with respect to the exact properties of the objects. There is a trade between precision for speed, generality, and the ability to make predictions that prove to be ``good enough'' for the everyday activities of human beings, hence that minimize Free Energy. In this type of scenario, the ''preference'' would be in the form of a computational constraint.

\subsection{Symmetry in the brain}

It is historically known that through visual processing the brain discards information about identity preserving transformations of objects~\cite{fukushima1980,poggio2004}. While at the V1 layer of the visual cortex, information is encoded in high-dimensional ``entangled'' representations, transformation information is lost and results in ``exemplar'' neurons~\cite{Chang2017}. Each ``exemplar'' neuron fires accordingly to a specific sensory identity (e.g. a specific object), invariant to the object pose.

Only recently, along with the advancement in ML, a new point of view has become more popular. This new take stems from the equivariant representation of brain functionality. In the invariant representation, a transformed input is mapped to the same intermediate representation. In the equivariant representation, the transformation to the input observation is preserved and cascaded into the intermediate space. Equivariant representations have been highly adopted in the context of ML \cite{bVAE,steenbrugge2018,ferraro2022,disentangled}, and are considered crucial to achieving generalization features from the model at hand. The adoption of equivariant architectures is favored in the research of "disentangled" systems by the ML community where they have demonstrated their efficiency in generalization, imagination, and abstraction reasoning. 

Considering now an object as a unique composition of features. The equivariant representation exploits the fact that a certain subset of features may be invariant to certain transformations, but the overall representation is still likely to preserve information about the applied transformation. As an example, consider a red-colored mug, when applying a rotational transformation, the color feature is independent of it, instead features like pose, and lighting conditions will have a dependency on it. With respect to the disentangling property of the brain, from the neuroscience literature, we have proof that going along the visual cortex, many primary IT neurons have specific activation with respect to some of the generative factors they are exposed to\cite{stankiewicz2002,Chang2017,ARGUIN20003099}. 

\subsection{Symmetry versus disentanglement}

The type of experimentation performed in this work is similar to the one performed in the disentanglement work by Higgins et al. \cite{bVAE}, where they showed how higher $\beta$ results in better disentanglement in the latent space. The disentanglement effect is obtained by the use of an isotropic unit Gaussian $\mathcal{N}(0,I)$ as a fixed prior. This specific distribution enforces disentanglement by exploiting the isotropic nature of the distribution (i.e. diagonal covariance matrix).

In our architecture, however, the prior is learned, and our latent state spaces do not show disentanglement features, even for high $\beta$ factors. As future work, we could also experiment with extra regularization terms to encourage disentanglement, and investigate how this relates to model complexity and symmetries.

\section{Conclusion}
In this paper, we analyzed the relation between model complexity and symmetry exploitation in the context of an object-centric deep active inference framework. We first showed how lower model complexity leads to an increase in exploiting symmetries in the learned latent state space. Second, we investigated in more detail how the learned symmetries in latent space capture the physical symmetries of the object modeled. Finally, we demonstrated how lower complexity models can be exploited for inferring preference realizing actions, which immediately generalize to symmetric configurations. In future work, we aim to investigate further how agents can learn the optimal accuracy-complexity trade-off, in view of an agent that needs to realize a certain set of preferences. In this setup, instead of manually varying $\beta$ in the loss function, the agent should ideally converge to the least complex model that is able to realize the agent's preferences.
In their work Falorsi et al. \cite{falorsi0218}, proposed an ad-hoc reparameterization trick for distributions on the SO(3) group of rotations in 3D. Matching the topology of the latent data manifold to the one of the latent space. As an extension to the current work, we aim to investigate the impact of the optimizations introduced by Falorsi et al. in the symmetry exploitation scenario.

\section*{Author Contributions}
SF conceived and performed the experiments.
SF, TVa, and TVe contributed to the software implementation. SF and TVe took care of the writing of the manuscript. TVa, TVe and BD contributed to the review of the manuscript. BD supervised the whole project. All authors contributed to the article and approved the submitted version.

\section*{Funding}
This research received funding from the Flemish Government
(AI Research Program).

\section*{Data Accessibility} Data are available as supplementary material. Training dataset, trained models along with the code (for training routine and figures generation) are provided. The authors remain available for any clarification on its use. 

\bibliographystyle{splncs04}
\bibliography{references}

\clearpage
\appendix 

\end{document}